\documentclass[a4paper]{article}

\usepackage{INTERSPEECH2019}
\usepackage{algorithm}
\usepackage[noend]{algpseudocode}
\def\NoNumber#1{{\def\alglinenumber##1{}\State #1}\addtocounter{ALG@line}{-1}}
\usepackage{hyperref}

\title{Universal Adversarial Perturbations for Speech Recognition Systems}
\name{\text{*}Paarth Neekhara$^{1}$, \text{*}Shehzeen Hussain$^{2}$, Prakhar Pandey$^{1}$, Shlomo Dubnov$^{3}$, Julian McAuley$^1$, Farinaz Koushanfar$^{2}$ }
\address{
  $^1$UC San Diego Department of Computer Science\\
  $^2$UC San Diego Department of Electrical and Computer Engineering\\
  $^3$UC San Diego Department of Music\\
  \text{*} Equal contribution}
\email{pneekhar@ucsd.edu, ssh028@ucsd.edu}

\begin{document}

\maketitle
\begin{abstract}
In this work, we demonstrate the existence of universal adversarial audio perturbations that cause mis-transcription of audio signals by automatic speech recognition (ASR) systems. We propose an algorithm to find a single quasi-imperceptible perturbation, which when added to any arbitrary speech signal, will most likely fool the victim speech recognition model. Our experiments demonstrate the application of our proposed technique by crafting audio-agnostic universal perturbations for the state-of-the-art ASR system -- Mozilla DeepSpeech. Additionally, we show that such perturbations generalize to a significant extent across models that are not available during training, by performing a transferability test on a WaveNet based ASR system.
\end{abstract}

\section{Introduction}

Machine learning agents serve as the backbone of several speech recognition systems, widely used in personal assistants of smartphones and home electronic devices (e.g.~Apple Siri, Google Assistant). Traditionally, Hidden Markov Models (HMMs)~\cite{baum1967,baum1970maximization,hmm1,ahadi1997combined,bahl1986maximum,Rabiner89-ATO} were used to model sequential data but with the advent of deep learning, state-of-the-art speech recognition systems are based on Deep Neural Networks (DNNs)~\cite{deepspeech2,parallelwavenet,wavenet,mozilladeepspeech}. 

However, several studies have demonstrated that DNNs are vulnerable to adversarial examples \cite{goodfellow6572explaining,obfuscated-gradients,Carlini2017TowardsET,atscale,limitations}. An adversarial example is a sample from the classifier's input domain which has been perturbed in a way that is intended to fool a victim machine learning (ML) model. While the perturbation is usually imperceptible, such an adversarial input can mislead neural network models deployed in real-world settings causing it to output an incorrect class label with higher confidence.  

A vast amount of past research in adversarial machine learning has shown such attacks to be successful in the image domain  ~\cite{intriguing,limitations, papernot1, papernot2,advpatch,harnessing}. However, few works have addressed attack scenarios involving other modalities such as audio. This limits our understanding of system vulnerabilities of many commercial speech recognition models employing DNNs, such as Amazon Alexa, Google Assistant, and home electronic devices like Amazon Echo and Google Home. Recent studies that have explored attacks on automatic speech recognition (ASR) systems \cite{asrblack,targetattacks,hidden, DBLP:journals/corr/abs-1810-11793}, have demonstrated that adversarial examples exist in the audio domain. The authors of \cite{targetattacks} proposed targeted attacks where an adversary designs a perturbation that can cause the original audio signal to be transcribed to any phrase desired by the adversary. However, calculating such perturbations requires the adversary to solve an optimization problem for each data-point they wish to mis-transcribe. This makes the attack in-applicable in real-time since the adversary would need to re-solve the data-dependent optimization problem from scratch for every new data-point.

Universal Adversarial Perturbations \cite{universal} have demonstrated that there exist universal \textit{image-agnostic} perturbations which when added to any image will cause the image to be mis-classified by a victim network with high probability. The existence of such perturbations poses a threat to machine learning models in real world settings since the adversary may simply add the same pre-computed universal perturbation to a new image and cause mis-classification.

\noindent\textbf{Contributions:} In this work, we seek to answer the question 
``Do universal adversarial perturbations exist for neural networks in audio domain?'' We demonstrate the existence of universal audio-agnostic perturbations that can fool DNN based ASR systems.%
\footnote{Sound Examples:
\href{universal-audio-perturbation.herokuapp.com}{universal-audio-perturbation.herokuapp.com} 
} 
We propose an algorithm to design such universal perturbations against a victim ASR model in the \textit{white-box setting}, where the adversary has access to the victim's model architecture and parameters. We validate the feasibility of our algorithm, by crafting such perturbations for Mozilla's open source implementation of the state-of-the-art speech recognition system  DeepSpeech ~\cite{mozilladeepspeech}. Additionally, we discover that the generated universal perturbation is transferable to a significant extent across different model architectures. Particularly, we demonstrate that a universal perturbation trained on DeepSpeech can cause significant transcription error on a WaveNet \cite{wavenet} based ASR model. 


\section{Related Work}
\noindent\textbf{Adversarial Attacks in the Audio Domain: } 
Adversarial attacks on ASR systems have primarily focused on \textit{targeted attacks} to embed carefully crafted perturbations into speech signals, such that the victim model transcribes the input audio into a specific malicious phrase, as desired by the adversary \cite{asrblack,targetattacks,mfccattack, hidden,usenixaudio}. Prior works \cite{hidden,usenixaudio} demonstrate successful attack algorithms targeting traditional speech recognition models based on HMMs and GMMs, that operate on Mel Frequency Cepstral Coefficient (MFCC) representation of audio. For example, in Hidden Voice Commands \cite{hidden}, the attacker uses inverse feature extraction to generate obfuscated audio that can be played over-the-air to attack ASR systems. However, obfuscated samples sound like random noise rather than normal human perceptible speech and therefore come at the cost of being fairly perceptible to human listeners. Additionally, these attack frameworks are not end-to-end, which render them impractical for studying the vulnerabilities of modern ASR systems -- that are entirely DNN based. 

In more recent work \cite{targetattacks}, Carlini \emph{et al.}~propose an end-to-end white-box attack technique to craft adversarial examples, which transcribe to a target phrase. Similar to the work in images, they propose a gradient-based optimization method that replaces the cross-entropy loss function used for classification, with a Connectionist Temporal Classification (CTC) loss~\cite{graves2006connectionist} which is optimized for time-sequences. The CTC-loss between the target phrase and the network's output is backpropagated through the victim neural network and the MFCC computation, to update the additive adversarial perturbation. The adversarial samples generated by this work are quasi-perceptible, motivating a separate work \cite{psychoacoustic} to minimize the perceptibility of the adversarial perturbations using psychoacoustic hiding. 

Designing adversarial perturbations using all the above mentioned approaches requires the adversary to solve a data dependent optimization problem for each input audio signal the adversary wishes to mis-transcribe, making them ineffective in a real-time attack scenario. In other words, targeted attacks must be customized for each segment of audio, a process that cannot yet be done in real-time. The existence of universal adversarial perturbations (described below) can pose a more serious threat to ASR systems in real-world settings since the adversary may simply add the same pre-computed universal adversarial perturbation to any input audio and fool the DNN based ASR system.

\noindent \textbf{Universal Adversarial Perturbations:} The authors of \cite{universal} craft a single universal perturbation vector which can fool a victim neural network to predict a false classification output on the majority of validation instances. Let $\hat{k}(x)$ be the classification output for an input $x$ that belongs to a distribution $\mu$. The goal is to find a perturbation $v$ such that:
$\hat{k}(x+v) \neq \hat{k}(x)$ for ``most'' $x \in \mu$. This is formulated as an optimization problem with constraints to ensure that the universal perturbation is within a specified p-norm and is also able to fool the desired number of instances in the training set. The proposed algorithm iteratively goes over the training dataset to build a universal perturbation vector that pushes each data point to its decision boundary. The authors demonstrate that it is possible to find a quasi-imperceptible universal perturbation that pushes most data points outside the correct classification region of a victim model. More interestingly, the work demonstrates that the universal perturbations are transferable across models with different architectures. The perturbation produced using one network such as VGG-16 can also be used to fool another network e.g. GoogLeNet showing that their method is doubly universal. \textit{Universal adversarial perturbations} for images focuses on the goal of mis-classification and cannot directly be applied to the more challenging goal of mis-transcription by Speech Recognition System. In our work we address this challenge and solve an alternate optimization problem to adapt the method for designing universal adversarial perturbations for ASR systems.

\section{Methodology}
\subsection{Threat Model}

\begin{figure}[htp]
    \centering
    \includegraphics[width=1.0\columnwidth]{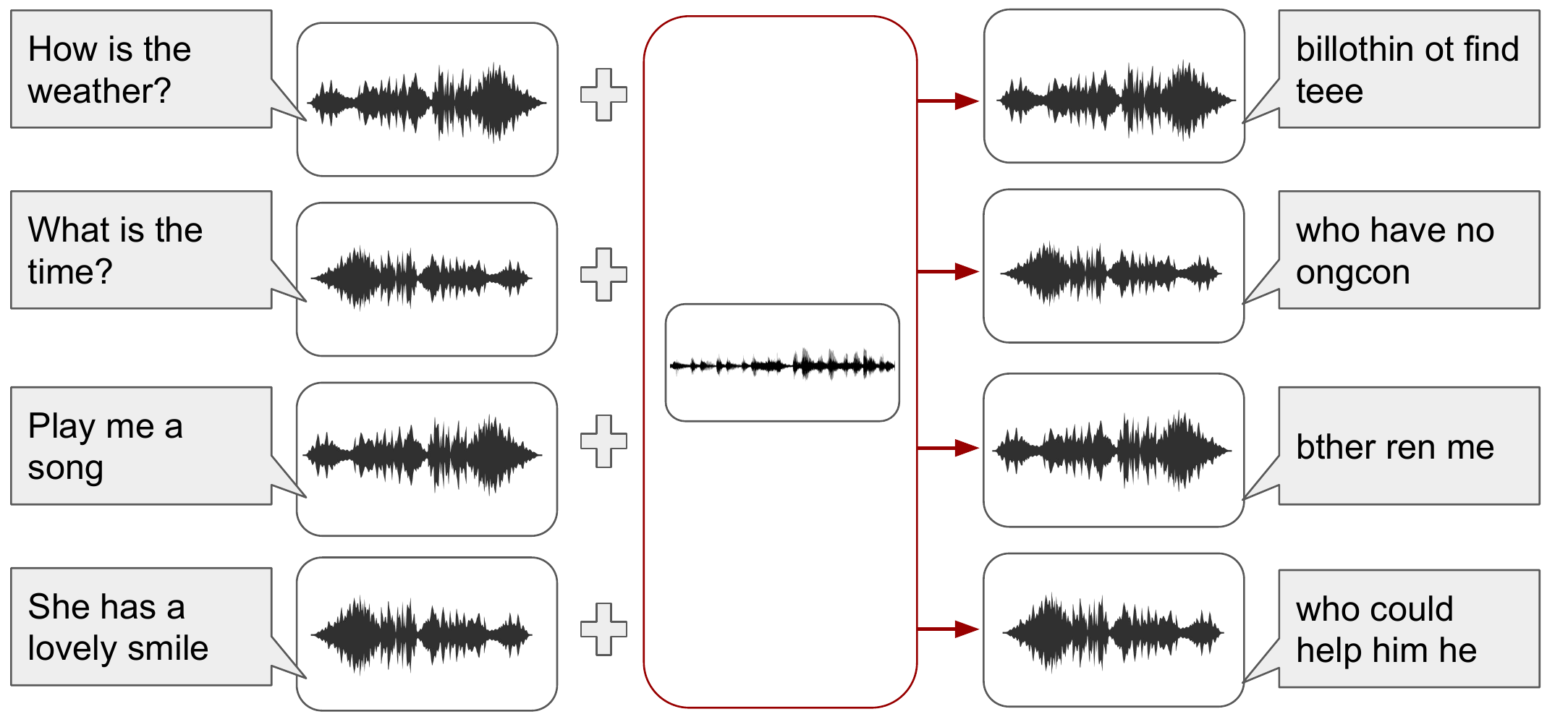}
    \caption{Threat Model: We aim to find a single perturbation which when added to any arbitrary audio signal, will most likely cause an error in transcription by a victim Speech Recognition System}
    \label{fig:my_label}
\end{figure}

\label{NED}
We aim to find a universal audio perturbation, which when added to any speech waveform, will cause an error in transcription by a speech recognition model with high probability. For the success of the attack, the error in the transcription should be high enough so that the transcription of the perturbed signal (adversarial transcription) is incomprehensible and the original transcription cannot be deduced from the adversarial transcription. As discussed in \cite{targetattacks}, the transcription \textit{``test sentence''} mis-spelled as \textit{``test sentense''} does little to help the adversary. To make the adversary's goal challenging, we report success only when the Character Error Rate (CER) or the normalized Levenshtein distance \textit{(Edit Distance)} \cite{editdistance} between the original and adversarial transcription is greater than a particular threshold. Formally, we define our threat model as follows:

Let $\mu$ denote a distribution of waveforms and $C$ be the victim speech recognition model that transcribes a waveform $x$ to $C(x)$. The goal of our work is to find perturbations $v$ such that:

$$\mathit{CER}( C(x), C(x + v)) > t \text{ for ``most'' } x \in \mu $$

Here, $\mathit{CER}(x, y)$ is the edit distance between the strings $x$ and $y$ normalized  \cite{editdistance} by the $length$ of $x$ i.e $$\mathit{CER}(x,y) = \frac{\mathit{EditDistance}(x, y)}{\mathit{length}(x)}$$

The threshold $t$ is chosen as 0.5 for our experiments i.e.,~we report success only when the original transcription has been \textit{edited} by at least $50\%$ of its length using character \textit{removal, insertion, or substitution} operations.

The universal perturbation signal $v$ is chosen to be of a fixed length and is cropped or zero-padded at the end to make it equal to length of the signal $x$. 

\subsection{Distortion Metric} \label{distmetric}
To quantify the distortion introduced by some adversarial perturbation $v$, an $l_\infty$ metric is commonly used in the space of images. Following the same convention, in the audio domain \cite{Carlini2017TowardsET}, the loudness of the perturbation can be quantified using the $\mathit{dB}$ scale, where $\mathit{dB}(v) = \max_i(20.\log_{10} (v_i)).$
We calculate $dB_x(v)$ to quantify the relative loudness of the universal perturbation $v$ with respect to an original waveform $x$ where: $$\mathit{dB}_x(v) = \mathit{dB}(v) - \mathit{dB}(x)$$
\noindent Since the perturbation introduced is quieter than the original signal, $\mathit{dB}_x(v)$ is a negative value, where smaller values indicate quieter distortions. In our results, we report the average relative loudness: $\mathit{dB}_x(v)$ across the whole test set to quantify the distortion introduced by our universal perturbation.

\subsection{Problem Formulation and Algorithm}
Our goal to find a quasi-imperceptible universal perturbation vector $v$ such that it mis-transcribes \textit{most} data points sampled from a distribution $\mu$. Mathematically, we want to find a perturbation vector $v$ that satisfies:
\begin{enumerate}
    \item $ \|v\|_\infty < \epsilon$
    \item $\underset{x \sim \mu}{P} \left( CER(C(x), C(x+v) > t) \right) \geq \delta.$\\
\end{enumerate}
Here $\epsilon$ is the maximum allowed $l_\infty$ norm of the perturbation, $\delta$ is the desired attack success rate and $t$ is the threshold CER chosen to define our success criteria. 

To solve the above problem, we adapt the universal adversarial perturbation algorithm proposed by \cite{universal} to find universal adversarial perturbations for the goal of \textit{mis-transciption} of speech waveforms instead of \textit{mis-classification} of data (images).  Let $X = {x_1, x_2, \ldots, x_m}$ be a set of speech signals sampled from the distribution $\mu$. Our Algorithm (\ref{algorithm1}) goes over the data-points in $X$ iteratively and gradually builds the perturbation vector $v$. At each iteration $i$, we seek a minimum perturbation $\Delta v_i$, that causes an error in the transcription of the current perturbed data point $x_i + v$. We then add this additional perturbation $\Delta v_i$ to the current universal perturbation $v$ and clip the new perturbation $v$, if necessary, to satisfy the constraint $ \|v\|_\infty < \epsilon$.

\begin{algorithm}
\caption{Universal Adversarial Perturbations for Speech Recognition Systems}\label{universaltrain}
\begin{algorithmic}[1]

\State \textbf{input:} Training Data Points $X$, Validation Data Points $X_v$ Victim Model $C$, allowed distortion level $\epsilon$, desired success rate $\delta$
\State \textbf{output:} Universal Adversarial Perturbation vector $v$
\State Initialize $v \gets 0$
\While {$\mathit{SuccessRate}(X_{v}) < \delta$}
\For {each data point $x_i \in X$ }
\If {$\mathit{CER}( C(x_i + v + r), C(x_i)) < t$}
\State {Compute min perturbation that }
\NoNumber {mis-transcribes $x_i + v$:}
\NoNumber{ \normalsize{$\Delta v_i \gets \arg\min_{r} \| r \|_2 \text{ s.t.:} $} }
\NoNumber {$\mathit{CER}( C(x_i + v + r), C(x_i)) > t$}
\State Update and clip universal perturbation $v$:
\NoNumber {$v = \mathit{Clip}_{v,\epsilon}(v + \Delta v_i)$}
\EndIf
\EndFor
\EndWhile
\end{algorithmic}
\label{algorithm1}
\end{algorithm}

At each iteration we need to solve the following optimization problem, that seeks a minimum (under \textit{l2} norm) additional perturbation  $\Delta v_i$, to mis-transcribe the current perturbed audio signal $x_i + v$:
\begin{equation}  \label{eqnew}
\Delta v_i \gets \arg\min_{r} \| r \|_2 \text{ s.t. } CER( C(x_i + v + r), C(x_i)) > t
\end{equation} 

It is non-trivial to solve the above optimization in its current form. In \cite{universal}, the authors try to solve a similar optimization problem for the goal of \textit{mis-classification} of data points. They approximate its solution using DeepFool \cite{deepfool} which finds a minimum perturbation vector that pushes a data point to its decision boundary. Since we are tackling a more challenging goal of \textit{mis-transcription} of signals where we have decision boundaries for each audio frame across the time axis, the same idea cannot be directly applied. Therefore, we approximate the solution to the optimization problem given by Equation \ref{eqnew} by solving a more tractable optimization problem:

\begin{equation}
\begin{split}
& \text{Minimize } J(r) \text{ where}\\
& J(r) = c\|r\|^2 + L(x_i+v+r, C(x_i)) \\
& \text{s.t.  }  \|v + r\|_\infty < \epsilon\\
& \text{where }L(x, y) =  -\mathit{CTCLoss}(f(x), y)
\end{split}
\label{eq2}
\end{equation}

In other words, to mis-transcribe the signal, we aim to maximize the CTC-Loss between the predicted probability distributions of the perturbed signal $f(x_i+v+r)$ and the original transcription $C(x_i)$ while having a regularization penalty on the $l\-2$ norm of $r$. Since this a non-convex optimization problem, we approximate its solution using iterative gradient sign method \cite{iterativeFGSM}:
\begin{equation} \label{eq3}
\begin{split}
& r_0 = \overrightarrow{0} \\
& r_{N+1} = \mathit{Clip}_{r+v,\epsilon} \{ r_N - \alpha 
\mathit{sign}( \Delta_{r_N} J(r_N) \}\\
\end{split}
\end{equation}

Note that the error $J$ is back-propagated through the entire neural network and the MFCC computation to the perturbation vector $r$. We iterate until we reach the desired CER threshold $t$ for a particular data point $x_i$. The regularization constant $c$ is chosen through hyper-parameter search on a validation set to find the maximum success rate for a given magnitude of allowed perturbation.

\section{Experimental Details}
We demonstrate the application of our proposed attack algorithm on the pre-trained \textit{Mozilla DeepSpeech} model \cite{mozilladeepspeechgit,mozilladeepspeech}. We train our algorithm on the Mozilla Common Voice Dataset \cite{mozilladeepspeech} which contains 582 hours of audio across 400,000 recordings in English. We train on a randomly selected set \textit{X} containing 5,000 audio files from the training set and evaluate our model on both the training set $X$ and the entire unseen validation set of the Mozilla Common Voice Dataset. We analyze the effect of the size of the set $X$ below.
The length of our universal adversarial perturbation is fixed to 150,000 samples which corresponds to around 9 seconds of audio at 16 KHz. The universal adversarial perturbations are trained using our proposed algorithm \ref{algorithm1} with a learning rate $\alpha = 5$ and the regularization parameter $c$ set to 0.5. 

\noindent\textbf{Evaluation:} We utilize two metrics: \textit{i) Mean CER} - Character Error Rate averaged over the entire test set and \textit{ii) Success Rate} to evaluate our universal adversarial perturbations. We report success on a particular waveform, if the \textit{CER} between the original and adversarial transcription (Section \ref{NED}) is greater than 0.5. The amount of perturbation is quantified using mean relative distortion $dB_x(v)$ over the test set (Refer to Section \ref{distmetric}).

\section{Results}
Table \ref{table:table2} shows the results of our algorithm for different allowed magnitude of universal adversarial perturbation on both the training set X and the unseen Test Set. Both the success rate and the Mean Character Error Rate (CER) increase with increase in the maximum allowed perturbation. We achieve a success rate of 89.06 \% on the validation set, with the mean distortion metric $dB_x(v) \approx -32 dB$. To interpret the results in context, $-32 dB$ is roughly the difference between ambient noise in a quiet room and a person talking \cite{dbnoise,targetattacks} . We encourage the reader to listen to our adversarial samples and their corresponding transcriptions on our web page (link in the footnote of the first page)

\begin{table}[ht]
\caption{Results of our algorithm for different allowed magnitude of universal adversarial perturbation}
\label{table:table2}
\resizebox{\columnwidth}{!}{%
\begin{tabular}{@{}c|ccc|ccc@{}}
\toprule
\multicolumn{1}{l|}{} & \multicolumn{3}{c|}{\textbf{Training Set (X)}} & \multicolumn{3}{c}{\textbf{Test Set}} \\ \midrule
\textbf{$ \mathbf{ \|v\|_\infty}$} & \textbf{\begin{tabular}[c]{@{}c@{}}Mean \\ $\mathbf{dB_x(v)}$\end{tabular}} & \textbf{\begin{tabular}[c]{@{}c@{}}Success \\ Rate (\%)\end{tabular}} & \textbf{\begin{tabular}[c]{@{}c@{}}Mean\\ CER\end{tabular}} & \textbf{\begin{tabular}[c]{@{}c@{}}Mean \\ $\mathbf{dB_x(v)}$\end{tabular}} & \textbf{\begin{tabular}[c]{@{}c@{}}Success \\ Rate (\%)\end{tabular}} & \textbf{\begin{tabular}[c]{@{}c@{}}Mean\\ CER\end{tabular}} \\ 
\midrule
100 & -42.03 & 57.46 & 0.63 & -41.86 & 56.13 & 0.64 \\ 
150 & -38.51 & 72.78 & 0.81 & -38.34 & 72.49 & 0.82 \\ 
200 & -36.01 & 83.27 & 0.92 & -35.84 & 80.47 & 0.95 \\ 
300 & -32.49 & 89.52 & 1.10 & -32.32 & 89.06 & 1.11 \\ 
400 & -30.18 & 90.60 & 1.06 & -29.82 & 88.24 & 1.07 \\ 
\bottomrule
\end{tabular}%
}
\end{table}

\begin{figure}[htp]
    \centering
    \includegraphics[width=0.9\columnwidth]{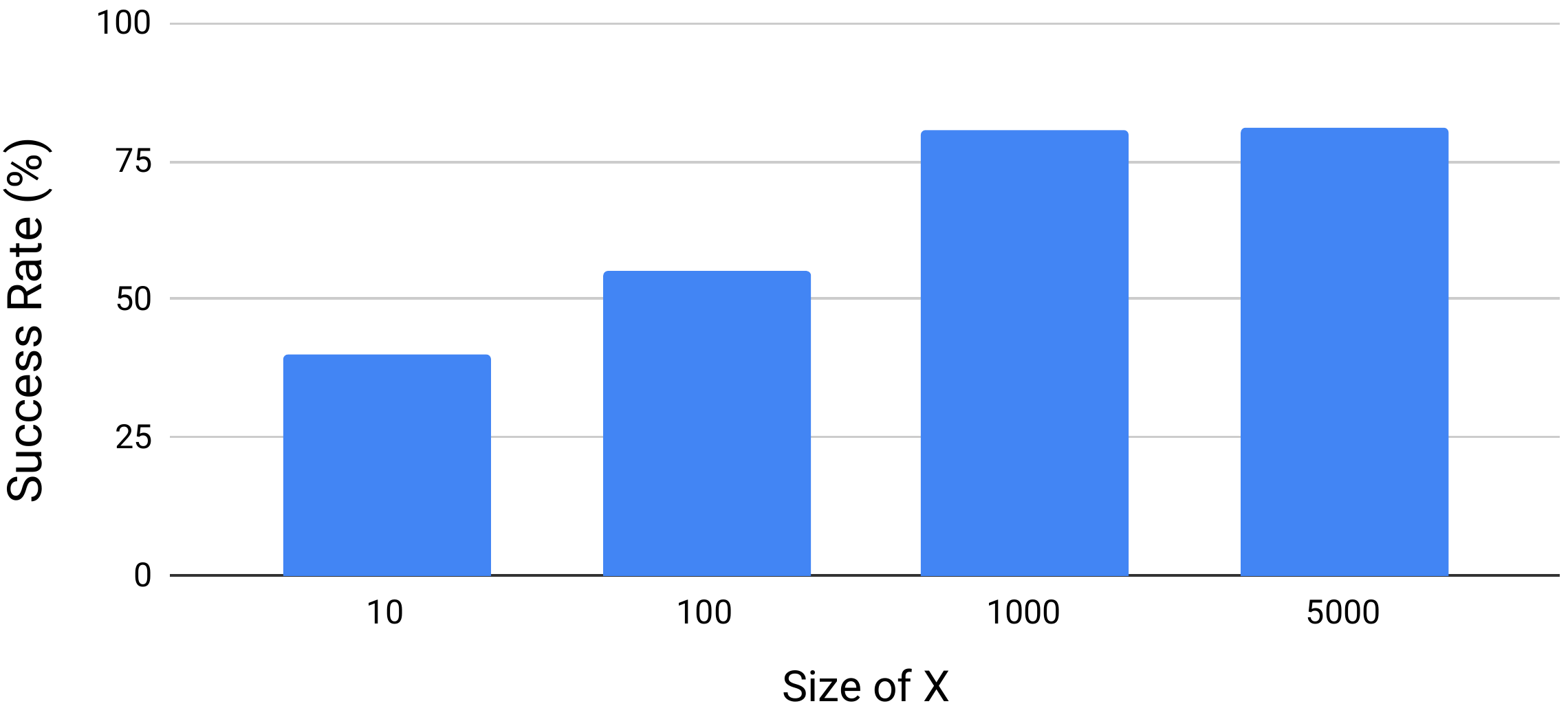}
    \caption{Attack Success Rate on the test set vs.~the number of audio files in the training set X}
    \label{fig:sizechart}
\end{figure}

\noindent Figure \ref{fig:sizechart} shows the success rate and mean edit distance compared to the size of the training set $X$ for maximum allowed perturbation $ \|v\|_\infty=200$ (Mean $dB_x(v) = -36.01$). We observe that it is possible to train our proposed algorithm on very few examples and achieve reasonable success rates on unseen data. For example, training on just 1000 examples can achieve a success rate of 80.47 \% on the test set. 

\subsection{Effectiveness of universal perturbations}
\begin{figure}[htp]
    \centering
    \includegraphics[width=0.9\columnwidth]{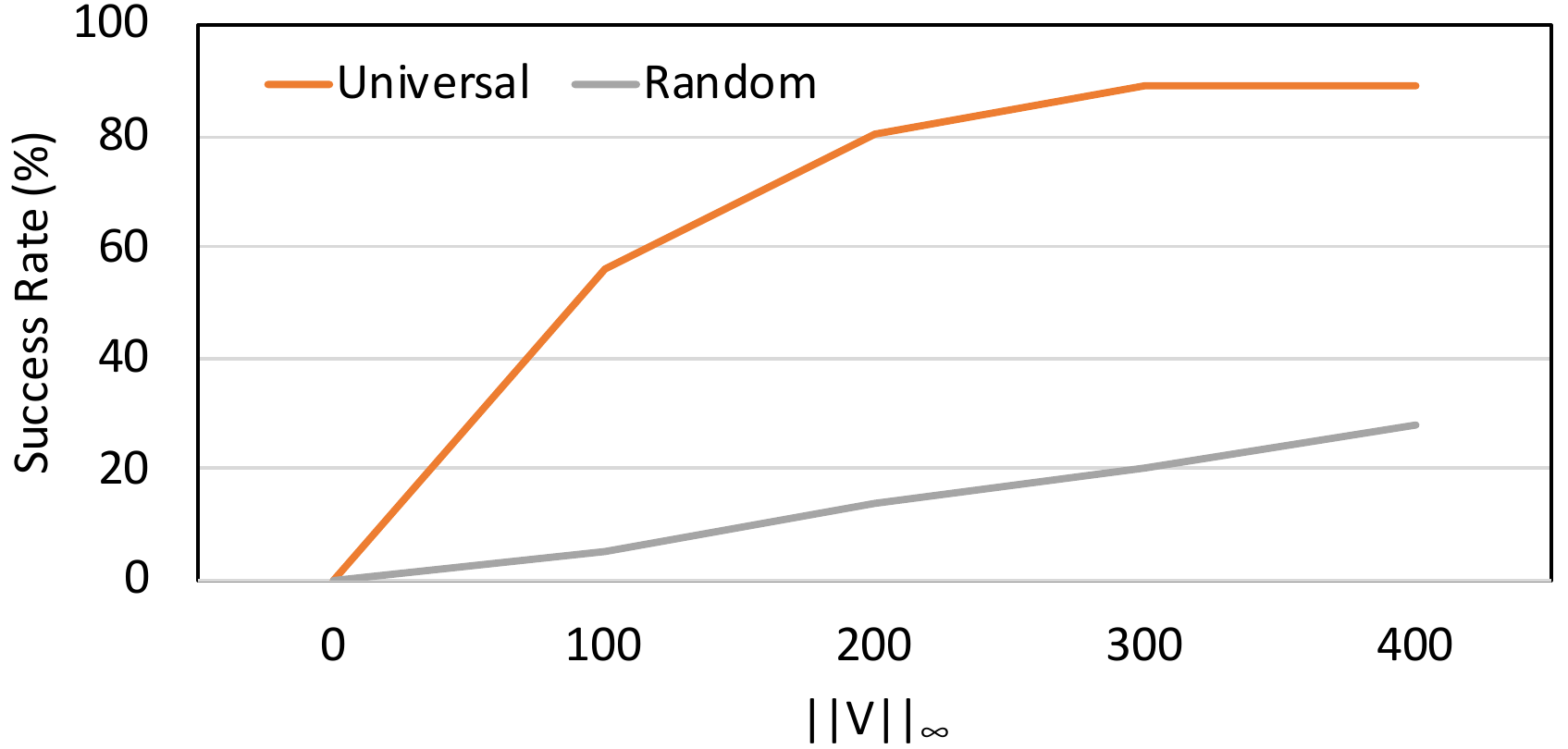}
    \caption{Success Rate vs $ \|v\|_\infty$ of universal and random perturbations.}
    \label{fig:baseline}
\end{figure}

In order to assess the vulnerability of the victim Speech Recognition System to our attack algorithm, we compare our universal perturbation with random (uniform) perturbation having the same magnitude of distortion (same $\|v\|_\infty$) as our universal adversarial perturbation. Figure \ref{fig:baseline} shows the plot of success rate vs.~the magnitude of the perturbation for each of these perturbations. It can be seen that universal adversarial perturbations are able to achieve high success rate with very low magnitude of distortion as compared to a random noise perturbation. For example, for allowed perturbation $ \|v\|_\infty = 100$ our universal perturbation achieves a success rate of $65\%$ which is substantially higher than the success rate of random noise. 
This implies that for the same magnitude of distortion, distorting an audio waveform in a random direction is significantly less likely to cause mis-transcription as compared to distorting the waveform in the direction of universal perturbation. Our results support the hypothesis discussed in \cite{universal}, demonstrating that universal adversarial perturbations exploit geometric correlations in the decision boundaries of the victim model.


\begin{table}[htp]
\centering
\caption{Results of the same universal adversarial perturbation on two victim models: Wavenet and Mozilla DeepSpeech. The universal perturbation was trained on the DeepSpeech model.}
\label{transfertable}
\resizebox{0.9\columnwidth}{!}{%
\begin{tabular}{@{}cc|cc|cc@{}}
\toprule
\multicolumn{2}{r|}{\textbf{}} & \multicolumn{2}{c|}{\textbf{Wavenet}} & \multicolumn{2}{c}{\textbf{Mozilla DeepSpeech}} \\ \midrule
\textbf{$ \mathbf{ \|v\|_\infty}$} & \textbf{\begin{tabular}[c]{@{}c@{}}Mean \\ dBx(v)\end{tabular}} & \textbf{\begin{tabular}[c]{@{}c@{}}Success \\ Rate (\%)\end{tabular}} & \textbf{\begin{tabular}[c]{@{}c@{}}Mean \\ CER\end{tabular}} & \textbf{\begin{tabular}[c]{@{}c@{}}Success \\ Rate (\%)\end{tabular}} & \textbf{\begin{tabular}[c]{@{}c@{}}Mean \\ CER\end{tabular}} \\ 
\midrule
150 & -38.34 & \textbf{26.97} & \textbf{0.37} & 72.49 & 0.82 \\ 
200 & -35.84 & \textbf{31.18} & \textbf{0.40} & 80.47 & 0.95 \\ 
300 & -32.32 & \textbf{42.05} & \textbf{0.47} & 89.06 & 1.11 \\ 
400 & -29.82 & \textbf{63.28} & \textbf{0.60} & 88.24 & 1.07 \\ 
\bottomrule
\end{tabular}%
}
\end{table}

\subsection{Cross-model Transferability}

We perform a study on the transferability of adversarial samples to deceive ML models that have not been used for training the universal adversarial perturbation, i.e.,~their parameters and network structures are not revealed to the attacker. We train universal adversarial perturbations for Mozilla DeepSpeech and evaluate the extent to which they are valid for a different ASR architecture based on WaveNet~\cite{wavenet}. For this study, we use a publicly available pre-trained model of WaveNet~\cite{asrwavenet} and evaluate the transcriptions obtained using clean and adversarial audio for the same unseen validation dataset as used in our previous experiments. Our results in Table~\ref{transfertable} indicate that our attack is transferable to a significant extent for this particular setting. Specifically, when the mean $\mathit{dB}_x(v)=-29.82$, we are able to achieve a 63.28\% success rate while attacking the WaveNet based ASR model. This result demonstrates the practicality of such adversarial perturbations, since they are able to generalize well across data points and architectures.

\section{Conclusion}
In this work, we demonstrate the existence of audio-agnostic adversarial perturbations for speech recognition systems. We demonstrate that our audio-agnostic adversarial perturbation generalizes well across unseen data points and to some extent across unseen networks. 
Our proposed end-to-end approach can be used to further understand the vulnerabilities and blind spots of deep neural network based ASR system, and provide insights for building more robust neural networks.

\bibliographystyle{IEEEtran}

\bibliography{mybib}


\end{document}